\documentclass[10pt,twocolumn,letterpaper]{article}

\usepackage[pagenumbers]{cvpr} 

\newcommand{\CUT}[1]{}

\usepackage{color,xcolor}
\usepackage{multirow} 
\usepackage{booktabs} 
\usepackage{colortbl}
\definecolor{mygray}{gray}{.9}

\usepackage{pifont}
\usepackage{makecell}
\usepackage{graphicx}
\usepackage{amsmath}
\usepackage{booktabs}
\usepackage{wrapfig}
\usepackage{multirow}  
\usepackage{bm} \bm{}
\usepackage[linesnumbered,ruled,vlined]{algorithm2e}

\def\z{{\bm z}}

\def\X{{\mathbf X}}

\def\Y{{\mathbf Y}}
\def\A{{\mathbf A}}
\def\D{{\mathbf D}}

\def\Q{{\mathbf Q}}
\def\K{{\mathbf K}}
\def\V{{\mathbf V}}

\def\M{{\mathbf M}}

\definecolor{cvprblue}{rgb}{0.21,0.49,0.74}
\usepackage[pagebackref,breaklinks,colorlinks,allcolors=cvprblue]{hyperref}

\def\paperID{*****} 
\def\confName{CVPR}
\def\confYear{2026}

\title{FREE-Edit: Using Editing-aware Injection in Rectified Flow Models for Zero-shot Image-Driven Video Editing}

\author{  \\
Maomao Li${^{1}}$,\quad Yunfei Liu${^{2}}$,\quad Yu Li ${^{2}}$\\
[5pt]
${^1}$The University of Hong Kong \qquad
${^2}$International Digital Economy Academy (IDEA) \qquad \\
{\tt\small limaomao07@connect.hku.hk \qquad \{liuyunfei,\ liyu\}@idea.edu.cn} \\  
}

\begin{document}

\twocolumn[{%
\renewcommand\twocolumn[1][]{#1}%
\maketitle
\begin{center}
    \centering
    \captionsetup{type=figure}
\includegraphics[width=1\textwidth]{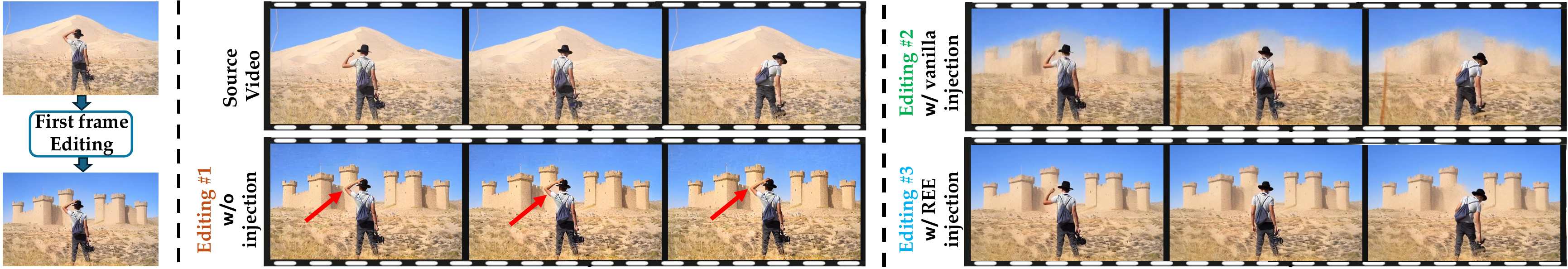}
\caption{\small We propose an \textbf{E}diting-awa\textbf{RE} (REE) feature injection method for pre-trained rectified \textbf{F}low models to conduct zero-shot image-driven video editing, dubbed FREE-Edit. Given the edited first frame, our method generates the output video, which propagates the editing contents while leaving other parts (\eg, motion) unchanged. The commonly used vanilla feature injection always leads to semantics conflicting with the edited first frame. We also provide the \textit{w/o} injection results, which cannot preserve motion well. In contrast, our REE feature injection method adaptively modulates the injection intensity of each token, achieving the most elegant results. }
    \label{fig:teaser}
\end{center}%
}]

\begin{abstract}
Image-driven video editing aims to propagate edit contents from the modified first frame to the remaining frames. Existing methods usually invert the source video to noise using a pre-trained image-to-video (I2V) model and then guide the sampling process using the edited first frame. Generally, a popular choice for maintaining motion and layout from the source video is intervening in the denoising process by injecting attention during reconstruction. However, such injection often leads to unsatisfactory results, where excessive injection leads to conflicting semantics with the source video while insufficient injection brings limited source representation. Recognizing this, we propose an \textbf{E}diting-awa\textbf{RE} (REE) injection method to modulate the injection intensity of each token. Specifically, we first compute the pixel difference between the source and edited first frame to form a corresponding editing mask. Next, we track the editing area throughout the entire video by using optical flow to warp the first-frame mask. Then, editing-aware feature injection intensity for each token is generated accordingly, where injection is not conducted in  editing areas. Building upon REE injection, we further propose a zero-shot image-driven video editing framework with recent-emerging rectified-\textbf{F}low models, dubbed FREE-Edit. Without fine-tuning or training, our FREE-Edit demonstrates effectiveness in various image-driven video editing scenarios, showing its capability to produce higher-quality outputs compared with existing techniques. Project page: \url{https://free-edit.github.io/page/}.
\end{abstract}

\section{Introduction}
\label{sec:intro}
In recent years, fruitful endeavours in diffusion models~\cite{ddim,ddpm} and rectified flow (RF)~\cite{liu2022flow,lipman2022flow} have been pursued, bringing a wide range of applications including text-to-image (T2I)~\cite{ldm,flux}, text-to-video (T2V)~\cite{blattmann2023align,yang2024cogvideox,hacohen2024ltx,wanx}, and image-to-video (I2V)~\cite{svd,yang2024cogvideox,hacohen2024ltx}. These models provide a pedestal for video editing, which requires that the generations meet editing requirements while keeping the non-editing area unchanged.

The current progress of video editing mainly lies in text-driven setting~\cite{fatezero,tokenflow,li2024video,wang2024taming}. However, to embrace more specific or localized editing, some approaches~\cite{videoshop,anyv2v,i2vedit} propose image-driven video editing, which first performs image editing on the first frame with any image editing tool (\eg, Photoshop) and then propagates the modification to the rest frames. This means the editing requirements can be specified using a reference image rather than text instructions~\cite{li2025qffusion}. Technically speaking, the current image-driven video editing methods follow the ``inversion-then-editing'' pipeline with an I2V model (\eg, SVD~\cite{svd}). Here, the source video is inverted into noise, which would be denoised later with the edited first frame as the conditional signal. Besides, since there is only limited motion representation in such a pure ``inversion-then-editing'' pipeline, they leverage source-video representation additionally by injecting reconstruction internal representations (\eg, Query or Key) to the editing denoising process.

Notwithstanding promising results, we argue that RF models have emerged as more powerful models than diffusion models on I2V, but the feasibility of using RF models for image-driven video editing is rarely explored. To make up for this gap, we first reproduce the aforementioned vanilla injection method with a pre-trained RF I2V model~\cite{hacohen2024ltx} under the ``inversion-then-editing'' pipeline, which replaces the model representation during editing with that during reconstruction. However, as seen in Fig.~\ref{fig:teaser}, it struggles to consistently maintain the appearance of the edited first frame. We argue that the unpleasing results of vanilla injection stem from its use of \textit{the same injection intensity for each token}. Here, excessive feature injection would conflict with those modified areas in the first frame, leading to undesired content leakage. Conversely, insufficient injection would bring limited source representations. Thus, vanilla injection is a sub-optimal solution.


Given the above considerations, we propose an \textbf{E}diting-awa\textbf{RE} (REE) injection method, which modulates the injection intensity of each token according to whether it is edited or not. Specifically, we design a modulation weight $\bm{\lambda}$ to implement a linear interpolation with Query $\Q$ (or Key $\K$) in the self-attention of the reconstruction branch and the corresponding $\tilde{\Q}$ (or $\tilde{\K}$) of the editing branch.
To compute the modulation weight $\bm{\lambda}$, we first use the difference between the source and edited first frames to mask the editing area. Next, we track the editing area throughout the entire video by evaluating optical flow~\cite{raft} for the source video, which is used to warp the first-frame mask. Then, the modulation weight $\bm{\lambda}$ can be obtained by inverting the warped mask sequence, where no injection is performed in the editing area to maintain appearance modification.


We apply the proposed REE injection method into the pre-trained \textbf{F}low model LTX-Video~\cite{hacohen2024ltx} to form a training-free image-driven video editing framework, which inspires the name of our method: FREE-Edit. Though simple, the extensive evaluation of FREE-Edit demonstrates our capability of generating high-quality appearance- and temporally consistent outputs. 
For example, as seen in Fig.~\ref{fig:teaser}, motion cannot be preserved well in FREE-Edit \textit{w/o}-injection due to the limited motion signal.
Besides, FREE-Edit \textit{w/} vanilla feature injection always leads to conflicting semantics from the source video with the edited appearance. In contrast, the proposed REE injection yields the most visually consistent results. To sum up, our contributions are:
\begin{itemize}
\item 
We propose a training-free image-driven video editing framework FREE-Edit, which equips an \textbf{E}diting-awa\textbf{RE} (REE) feature injection to a pre-trained \textbf{F}low model.
\item
Our REE injection designs a modulation weight $\bm{\lambda}$, which uses optical flow to track the edited mask of the first frame and does not perform injection in editing areas.
\item
Extensive experiments demonstrate that our FREE-Edit yields more elegant editing results in various scenarios than the current state-of-the-art editing methods.
\end{itemize}

\section{Related Works}
\noindent{\textbf{Text-driven Video Generation and Editing.}}
Early Text-to-Video (T2V) models ~\cite{an2023latent,blattmann2023align} opt to insert temporal layers a pretrained Text-to-Image (T2I) model~\cite{ldm} to obtain temporal modeling. Another line of methods~\cite{hacohen2024ltx,yang2024cogvideox} employs 3D VAE to incorporate 3D convolutions, which operate videos both spatially and temporally. Besides, text-driven video editing expects to generate an edited video that adheres to the target prompt while the original motion and semantic layout are preserved. Many efforts~\cite{tune,fatezero,tokenflow,li2024video,li2024vidtome} expand the U-Net architecture of a pre-trained T2I diffusion model (\eg, SD~\cite{ldm}) along the temporal axis for 3D correspondence, where the denoising process is performed under the target-prompt guidance. Recent progress~\cite{wang2024taming} has also included RF models to perform editing. Additionally, some methods use multiple frames~\cite{jain2024video,wang2024generative,guo2023sparsectrl,yang2024vibidsampler} or incorporate motion guidance~\cite{wang2024videocomposer,shi2024motion} for I2V generation.


\noindent{\textbf{Image-to-Video Generation and Editing.}}
Several representative works~\cite{svd,hacohen2024ltx,yang2024cogvideox} establish the foundation for image-to-video (I2V). Then, a group of methods~\cite{hacohen2024ltx,wanx} propose a first-frame conditioning model with rectified-flow training procedure.Besides, some techniques~\cite{anyv2v,videoshop,i2vedit} use these I2V models to carry out image-driven video editing by propagating the first edited frame. Here, besides training-free AnyV2V~\cite{anyv2v} and VideoShop~\cite{videoshop}, I2VEdit~\cite{i2vedit} performs finetuning on temporal attention layers for coarse motion extraction, which, however, is time- and resource-consuming.
For example, I2VEdit takes about 2 hours for a 40-frame video in an NVIDIA A100 GPU, which is unfavourable in real applications.
Moreover, very recent work Go-with-the-Flow~\cite{burgert2025go} proposes a motion-controlled method by fine-tuning video diffusion models, which also allows for first-frame guided video editing.


This paper follows the training-free trend and designs an \textbf{E}diting-awa\textbf{RE} (REE) feature injection method on a pre-trained flow I2V model. Specifically, we use optical flow to propagate the edited area in the first frame and then generate the modulation weight $\bm{\lambda}$ to ensure those pixels in the editing area are not subjected to improper feature injection.

\section{Preliminary} 
\noindent{\textbf{Rectified Flow Models.}}
Rectified Flow (RF)~\cite{liu2022flow,lipman2022flow,esser2024scaling} is a particular class of flow models that learn straight flow models, where the forward process corrupts the clean latent $\z_0$ linearly:
\begin{equation}
\begin{aligned}
\z_t = (1-t)\z_0 + t\z_1,
\end{aligned}
\label{rf_flow}
\end{equation}
where $t$ is sampled from a uniform distribution (\ie, $t \in \mathcal{U}(0,1)$). RF is trained to solve the least squares regression problem, which matches $v_{\theta}$ with $\z_1-\z_0$:
\begin{equation}
\begin{aligned}
\min_{\theta}\int_{0}^{1}\mathbb{E}[\parallel (\z_1-\z_0)-v_{\theta}(\z_t,t)\parallel^2]dt.
\end{aligned}
\label{loss}
\end{equation}
After the velocity field $v_{\theta}$ is trained, pure noise $\z_1$ is gradually recovered as a clean latent $\z_0$ using $N$ discrete timesteps $t_i$: 
\begin{equation}
\begin{aligned}
\z_{t_{i-1}} = \z_{t_i} + (t_{i-1}-t_i)v_{\theta}(\z_{t_i},t_i).
\end{aligned}
\label{rf_sample}
\end{equation}

\noindent{\textbf{Transformer Model for Image-to-Video.}} 
LTX-Video~\cite{hacohen2024ltx} is an open-source real-time generative model for I2V and T2V. Specifically, it uses a training procedure based on RF with 28 DiT transformer blocks~\cite{peebles2023scalable}.
Besides, it employs 3D causal VAEs to deal with spatio-temporal representation, where a $(1 + rn)$-frame video is compressed into $(1 + n)$ latent frames with temporal compression ratio $r$. Here, the first frame is encoded as a separate latent frame. Additionally, LTX-Video can perform real-time generation due to its highly compressed latent space.

\begin{figure*}[t]
  \centering
  \includegraphics[width=1\linewidth]{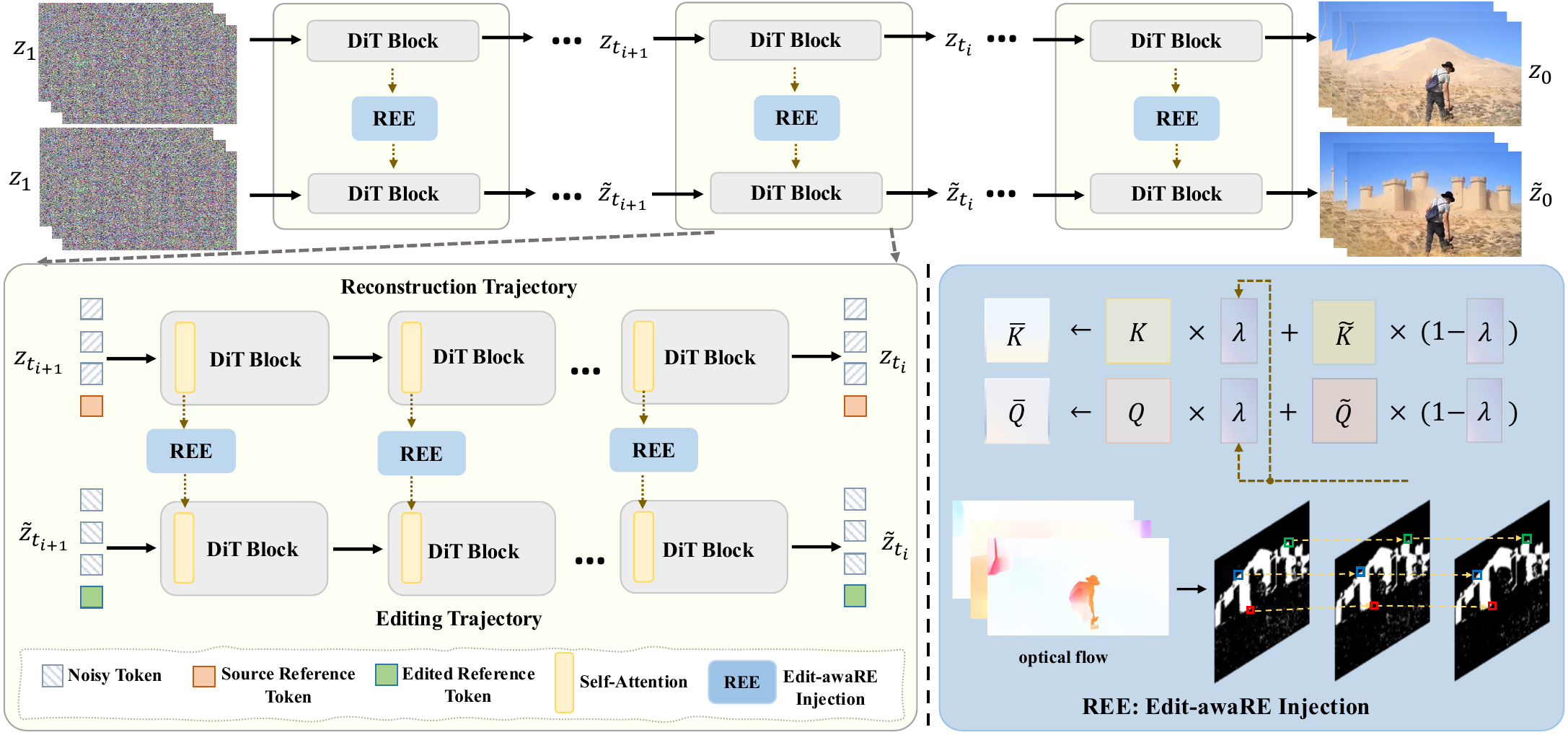}
  \caption{\small The pipeline illustration of our FREE-Edit. Top: It obeys an ``inversion-then-editing'' pipeline. Starting from inverted noisy latent $\z_1$, the reconstructed video $\z_0$ and edited video $\tilde{\z}_0$ take the source ${\X}^{1}$ and edited first frame $\hat{\X}^{1}$ as conditions, respectively. Bottom: We design an \textbf{E}diting-awa\textbf{RE} (REE) injection method, which designs a modulation weight $\bm{\lambda}$ to adaptively replace the intermediate model representations ($\tilde{\Q}$ and $\tilde{\K}$) in the editing process with those ($\Q$ and $\K$) in the reconstruction process through self-attention blocks. Here, we first use optical flow to warp the automatically calculated first-frame editing mask, which yields tracked editing masks for subsequent frames. Based on them, we compute the modulation weight $\bm{\lambda}$ for each token, where no injection is performed in editing regions. 
  }
  \label{fig:pipeline}
\end{figure*}

\section{Method}
\label{sec:method}
Given a source video $\X=[\X^0,\X^1...,\X^{rn}]$, image-driven video editing synthesizes the target video $\Y=[\Y^0,\Y^1,...,\Y^{rn}]$ that propagates the appearance of the first edit frame $\hat{\X}^1$ while remaining the rest parts unchanged (\eg, motion). Here, since temporal compression ratio $r$, we have ${\z}=[\z^0,\z^1...,\z^{n}]$.

The rest of this section is organized as follows. In Sec.~\ref{sec:3-1}, we first present a baseline editing framework that employs the vanilla feature injection on a pre-trained video RF model (\eg, LTX-Video~\cite{hacohen2024ltx}). Then, we elaborate on the proposed \textbf{E}diting-awa\textbf{RE} (REE) feature injection method and our FREE-Edit framework in Sec.~\ref{sec:3-2}.


\subsection{Vanilla Feature Injection on RF Models}
\label{sec:3-1}
Before introducing our FREE-Edit, we present a baseline framework for image-driven video editing, which follows the ``inversion-then-editing''~\cite{anyv2v,videoshop} pipeline using an RF I2V model LTX-Video~\cite{hacohen2024ltx}. The inversion process conducts sampling in the reverse direction. It maps the source video into a known latent space, which yields the latent noise at each time step $t$. To match RF models, we apply typical ODE RF inversion~\cite{rout2024semantic} and use the source first frame as condition:
\begin{equation}
\begin{aligned}
\z_{t_i} = {\rm RF\!-\!Inv}(v_{\theta}(\z_{t_{i-1}}, \X^1,{t_{i-1}}))
\end{aligned}
\label{rf_inversion}
\end{equation}
where ${\rm RF\!-\!Inv}$ stands for RF inversion. $\z_{t_i} \in \mathbb{R}^{(1+n) \times l \times c}$, where $l$ represents the length of token sequence and $c$ is dimension number. Then, the edited first frame $\hat{\X}^{1}$ can be used to guide the denoising process of the I2V model for generating the edited video $\Y$.

Considering motion signals from the inverted noise alone is limited, AnyV2V\cite{anyv2v} introduces a vanilla feature injection method in ``inversion-then-editing'' pipeline. Next, we will present how to reproduce it to match RF video models. Formally, we use $\A$ and $\tilde{\A}$ to denote the full \textit{spatiotemporal} self-attention maps in source denoising (reconstruction) and editing denoising trajectory, respectively. Then, to obtain more layout and motion representation from the source video, vanilla injection applies the attention maps ${\A}$ to replace attention maps $\tilde{\A}$, where $\tilde{\A}$ and $\A$ are obtained by:
\begin{equation}
\begin{aligned}
\tilde{\A}\!=\!softmax(\frac{\tilde{\Q}\tilde{\K}^{\top}}{\sqrt{d}}\cdot \tilde{\V}), {\A}\!=\!softmax(\frac{{\Q}{\K}^{\top}}{\sqrt{d}} \cdot {\V}),
\end{aligned}
\label{softmax}
\end{equation}
where $\Q$, $\K$, and $\V$ are the Query, Key, and Value embeddings in self-attention blocks of the source denoising (reconstruction) process while $\tilde{\Q}$, $\tilde{\K}$, and $\tilde{\V}$ are those of the editing denoising process. Besides, $\Q$, $\K$, $\tilde{\Q}$, and $\tilde{\K} \in \mathbb{R}^{(1+n) \times l \times d}$, where $d$ represents the dimension number of embeddings.

In practice, since \textit{spatiotemporal} self-attention maps ${\A}$ is parameterized by $\Q$ and $\K$, vanilla injection is applied to $\Q$ and $\K$ instead, yields the denoised latent $\tilde{\z}_{t_{i-1}}$:
\begin{equation}
\begin{aligned}
\tilde{\z}_{t_{i-1}} = v_{\theta}(\z_{t_{i}}, \hat{\X}^1,{t_{i}};\Q_{t_i}, \K_{t_i}),
\end{aligned}
\label{vanilla_injection}
\end{equation}
where $\tilde{\z}_{t_{i-1}} \in \mathbb{R}^{(1+n) \times l \times c}$, having the same size with $\z_{t_i}$.



\subsection{The Proposed REE Injection and FREE-Edit}
\label{sec:3-2}
\noindent{\textbf{Overview of Our REE Injection.}}
From Fig.~\ref{fig:teaser}, we can observe that vanilla injection always shows undesired content leakage, since excessive injection would conflict with the semantics of those modified areas in the first frame. However, insufficient injection often cannot bring enough motion signals. In view of the above dilemma, this paper proposes an \textbf{E}diting-awa\textbf{RE} (REE) feature injection method, which only performs injection in the non-editing area, thus preserving the editing content.

We notice that the vanilla injection uses the same injection intensity for each pixel or token even though the editing and non-editing areas are modified to different degrees. To deal with this, we propose to adaptively modulate the injection intensity of each token according to its editing degree. Formally, we define $\bm{\lambda} \in \mathbb{R}^{(1+n) \times l\times 1}$ as such a modulation weight, which is shared across channels.
Then, we can give a unified formulation of Query ($\bar{\Q}$) and Key ($\bar{\K}$) in our REE injection during the editing denoising process:
\begin{equation}
\begin{aligned}
\bar{\Q} = \bm{\lambda} \Q + (\mathbf{1}- \bm{\lambda}) \tilde{\Q}, \quad
\bar{\K} = \bm{\lambda} \K + (\mathbf{1}-\bm{\lambda}) \tilde{\K},
\end{aligned}
\label{general_inj}
\end{equation}
When $\bm{\lambda}=\mathbf{0}$, feature injection is disabled. When $\bm{\lambda}=\mathbf{1}$, our REE injection degenerates to the vanilla injection.

\noindent{\textbf{How to Calculate Modulation Weights.}} Generally speaking, considering the optical flow between two frames is a dense pixel displacement field that specifies the horizontal and vertical motion of each pixel, we use it to warp the first-frame editing mask, thus obtaining editing masks for subsequent frames. Then, with these masks, we calculate the modulation weight $\bm{\lambda}$ of all $(1+n)$ latent frames, where injection is only conducted in the non-editing area.


Specifically, we first calculate the difference map $\D^1$ between the source and the edited first frame as:
\begin{equation}
\begin{aligned}
\D^1 = \hat{\X}^1-\X^1,
\end{aligned}
\label{diff}
\end{equation}
where the dimension of each frame is $H \times W$. Then, we use a threshold $thr$ to map $\D^1$ into a binary mask $\M^1$:
\begin{equation}
\begin{aligned}
\M^1_j = \left\{\begin{matrix}
  255, & {\rm if} \ \D^1_j > thr \\ 
  0, &{\rm otherwise}.
\end{matrix}\right.
\end{aligned}
\label{map}
\end{equation} 
where $j \in \{0,1,H\times W -1\}$ is pixel index. Those locations with a value of 255 are regarded as editing pixels, while those locations with a value of 0 are non-editing pixels.

Next, we use RAFT~\cite{raft} to estimate optical flow of the source video, which is used to warp the first-frame mask $\M^1$ over the source video. Specifically, assuming $f^k$ is the optical flow between two consecutive frames $\X^{k}$ and $\X^{k+1}$, we map each coordinate $(x^{k},y^{k})$ in the $k$-th mask $\M^{k}$ to the coordinate $(x^{k+1},y^{k+1})$ in the $(k\!+\!1)$-th mask $\M^{k+1}$:
\begin{equation}
\begin{aligned}
x^{k+1} = x^{k}+f_x^{k}(x^{k}, y^{k}), \quad
y^{k+1} = y^{k}+f_y^{k}(x^{k}, y^{k}),
\end{aligned}
\label{autoregressive}
\end{equation}
where $k \in \{0,1,\ldots,rn\}$. Note that each new coordinate $(x^{k+1},y^{k+1})$ will be remapped to 0 or 255. When occlusion occurs, we set the corresponding value to 0 since we regard the occlusion area as a non-editing area.


\begin{figure*}[t]
  \centering
  \includegraphics[width=1\linewidth]{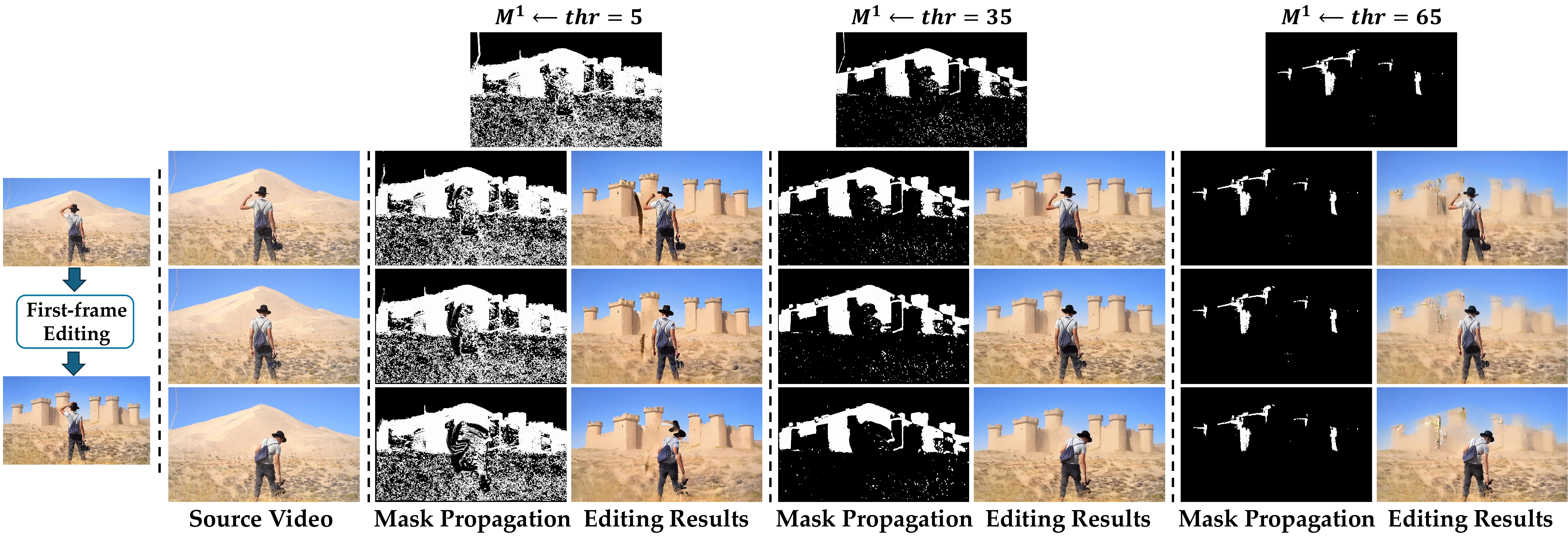}
   \vspace{-1.5em}
  \caption{ \small Qualitative ablation of different threshold ($thr$) in Eq.~(\ref{map}) for mask $\M^1$ generation, which are used to indicate editing regions.}
  \label{fig:ablation}
     \vspace{-0.6em}
\end{figure*}

The remaining problem is how to use the editing mask sequence $\M=\{\M^0,\M^1,...,\M^{rn}\}$ to generate modulation weight $\bm{\lambda}$. Since the temporal compression of the RF I2V model, we take the mask of the $\left \lceil \frac{r}{2} \right \rceil$-th frame for each $r$-frame video chunk. Thus, the \textbf{c}ompressed editing mask sequence $\M_{\mathbf{c}}=\{\M^0,\M^{\left \lceil \frac{r}{2} \right \rceil},...,\M^{\left \lceil \frac{r}{2}\right \rceil \cdot n}\}$ is with $(1+n)$ mask frames. Then, to match the dimension of Query and Key in Eq.~(\ref{general_inj}), we further \textbf{d}own-sample and flatten each mask frame in $\M_{\mathbf{c}}$ into size of $l\times 1$, yielding $\M_{c\cap d} \in \mathbb{R}^{(1+n) \times l \times 1}$.
Finally, our modulation weight $\bm{\lambda}$ can be obtained by inverting the $\M_{c\cap d}$:
\begin{equation}
\begin{aligned}
\bm{\lambda} = \mathbf{1} - \M_{c\cap d}.
\end{aligned}
\label{lambda}
\end{equation}
This means we only perform injection in non-editing areas.

\noindent{\textbf{FREE-Edit With Our REE Injection.}}
In Fig.~\ref{fig:pipeline}, we propose a training-free image-driven video editing framework, dubbed FREE-Edit, which obeys an ``inversion-then-editing'' pipeline. The core is our REE injection, which adaptively replaces $\tilde{\Q}$ and $\tilde{\K}$ in the editing trajectory with ${\Q}$ and ${\K}$ in the reconstruction editing trajectory through self-attention blocks in transformers. Here, clean tokens (i.e., source reference tokens and edited reference tokens corresponding to ${\X}^{1}$ and $\hat{\X}^{1}$) are combined with noisy tokens in each time step.

Note that text-driven technique Fatezero~\cite{fatezero} also explores attention blending with masks, which are generated from cross-attention maps of the text token associated with the foreground object. However, cross-attention maps struggle to indicate editing regions in certain editing types (\eg, adding objects), where the location of the added object is uncertain. In contrast, we use optical flow to track editing regions, which can be well adapted to various editing types.

Our method also differs from TokenFlow~\cite{tokenflow}, which propagates diffusion features during editing based on inter-frame correspondences calculated during inversion. Specifically, TokenFlow performs \textit{correspondence propagation} in text-driven editing, whereas FREE-Edit employs \textit{editing-mask propagation} in image-driven editing.

\section{Experiment}
\label{sec:exp}
\subsection{Experimental Setup}
\noindent{\textbf{{Implementation Details.}}
We conduct all experiments on a single Tesla A100 GPU using PyTorch~\cite{paszke2019pytorch}. We use the classifier-free guidance (CFG)~\cite{cfg} of 3 during the editing process, and that of 1 in the inversion process, which is under the same negative prompt \textit{“worst quality, inconsistent motion, blurry, jittery, distorted”}. Additionally, we adopt 50 time steps as noise schedule. We set $thr=35$ in Eq.~(\ref{map}) empirically. We use LTX-Video-2B~\cite{hacohen2024ltx} for the base I2V model for its real-time performance.
As for feature injection, we conduct Eq.~(\ref{general_inj}) in all transformer blocks and all time steps.

\noindent{\textbf{{I2V-Edit-Benchmark}.}} 
Considering there is no publicly available dataset for image-driven video editing, this paper constructs a mini benchmark named \textit{I2V-Edit-Bench} to evaluate the proposed method. Specifically, it contains 60 videos and their corresponding edited first frames.  The videos are collected from the Internet website, and DAVIS dataset~\cite{pont20172017}, and their edit frames are created by using professional software (e.g., Photoshop) and MagicBrush~\cite{magicbrush}. Each video consists of 41 to 200 frames. We provide 6 editing types: including \textit{object replacement, style transfer, adding/removing objects, changing attributes}, and \textit{background editing}.

\begin{table}[t]
\centering
\begin{tabular}{l|c|c}
\toprule  
Settings
&{\textit{CLIP score}}   $\uparrow$
&{\textit{Warp Error}}  $\downarrow$     
\\
\hline
$thr=5$ &0.920   &2.106  \\
$thr=35$  &\textbf{0.937} &\textbf{0.595}   \\
$thr=65$ &0.877  & 0.764 \\
\bottomrule
\end{tabular}
     \vspace{-0.6em}
\caption{\small Quantitative ablation of different threshold $thr$.}
\label{tab:ab_quanti}
\vspace{-0.3cm}
\end{table}


\noindent{\textbf{{Evaluation Metrics of Edited Videos}.}}
We use \textit{CLIP score} ($\uparrow$) to measure editing fidelity, which calculates the similarity of CLIP image embedding~\cite{clip} of each edited frame and the given first edit frame. Besides, we apply \textit{Warp Error} ($\downarrow$)~\cite{tokenflow,li2024video} to reflect temporal consistency. Specifically, we estimate optical flow~\cite{raft} of the input video to warp the edited frames and compute average MSE between warped frames and the target ones. Moreover, with our generated mask sequence $\M$, we measure the fidelity of the non-editing regions by five metrics: CLIP score$^{+}$ ($\uparrow$), Warp Error$^{+}$ ($\downarrow$), SSIM$^{+}$~\cite{SSIM} ($\uparrow$), PSNR$^{+}$~\cite{psnr} ($\uparrow$), and LPIPS$^{+}$ ($\downarrow$)~\cite{lpips}, respectively.

\begin{figure*}[t]
  \centering
  \includegraphics[width=1\linewidth]{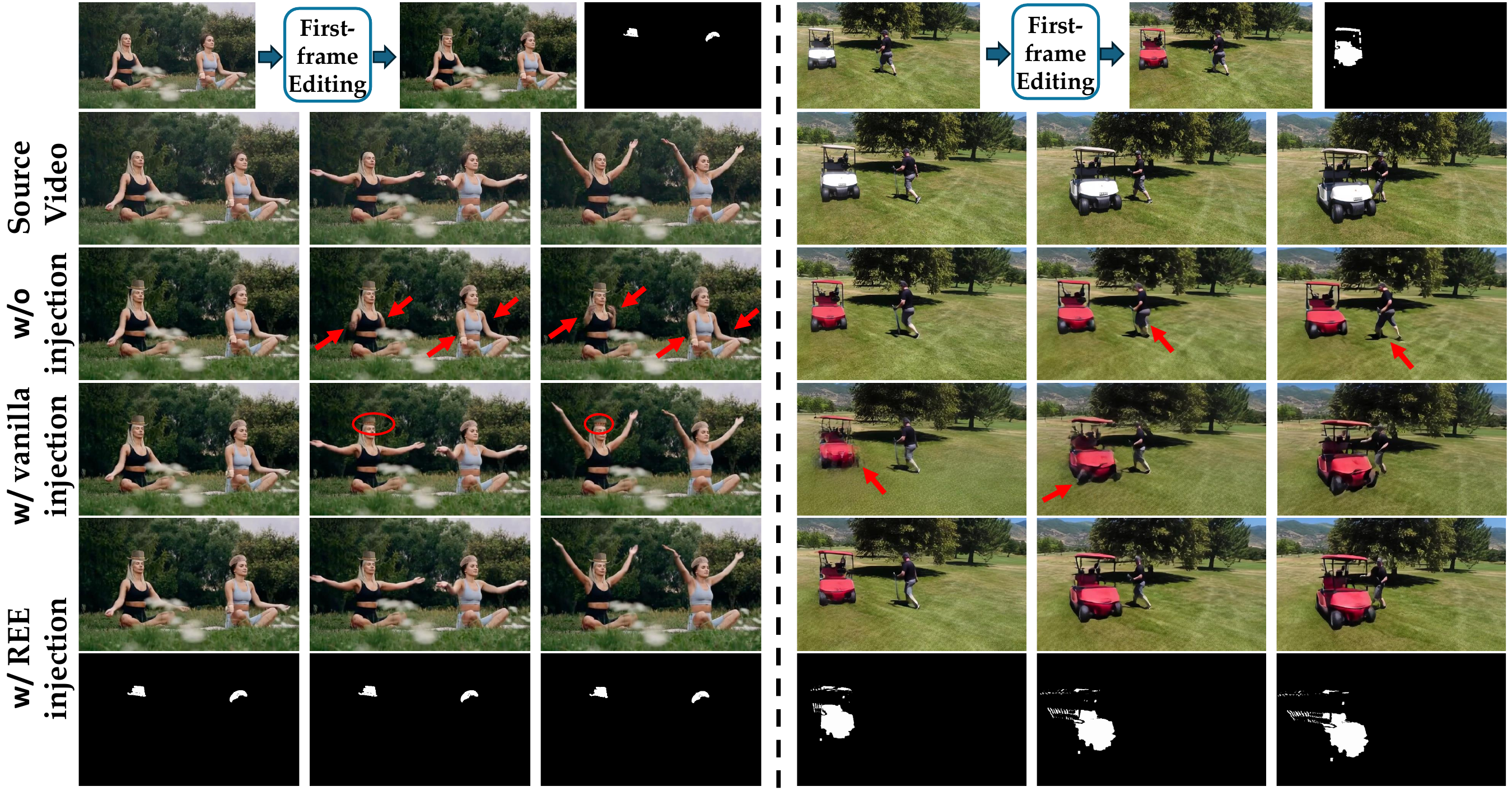}
  \caption{\small Qualitative comparison between standard FREE-Edit (\textit{w/} REE injection), and that of \textit{w/o} injection, and \textit{w/} vanilla injection.  }
  \label{fig:quali1}
\end{figure*}

\subsection{Ablation Study}
\label{sub:ablation_study}
We perform an ablation study to qualitatively and quantitatively validate the effect of threshold $thr$ in Eq.~(\ref{map}) for generating mask $\M^1$. In Fig.~\ref{fig:ablation}, we show the mask propagation and editing results when $thr$ is 5, 35, and 65, respectively. Here, $thr=5$ leads to a relatively large editing region in $\M^1$. Since our FREE-Edit does not perform feature injection in the editing region, the corresponding editing results of $thr=5$ still present unsatisfactory motion (see man's hat). In contrast, $thr=65$ yields a smaller editing region in $\M^1$, which brings excessive feature injection from the source video, thus suffering from undesired content leakage. In comparison, $thr=35$ obtains the most accurate editing regions and the best editing results. Thus, we use $thr=35$ as the default setting. Besides, Tab.~\ref{tab:ab_quanti} reports \textit{CLIP score} and \textit{Warp Error} of the editing results of different $thr$. Here, $thr=5$ performs poorly in temporal coherence, and $thr=65$ is inferior in edit consistency since they correspond to larger and smaller editing masks, leading to insufficient and excessive injection.

\begin{figure*}[t]
  \centering
  \includegraphics[width=0.88\linewidth]{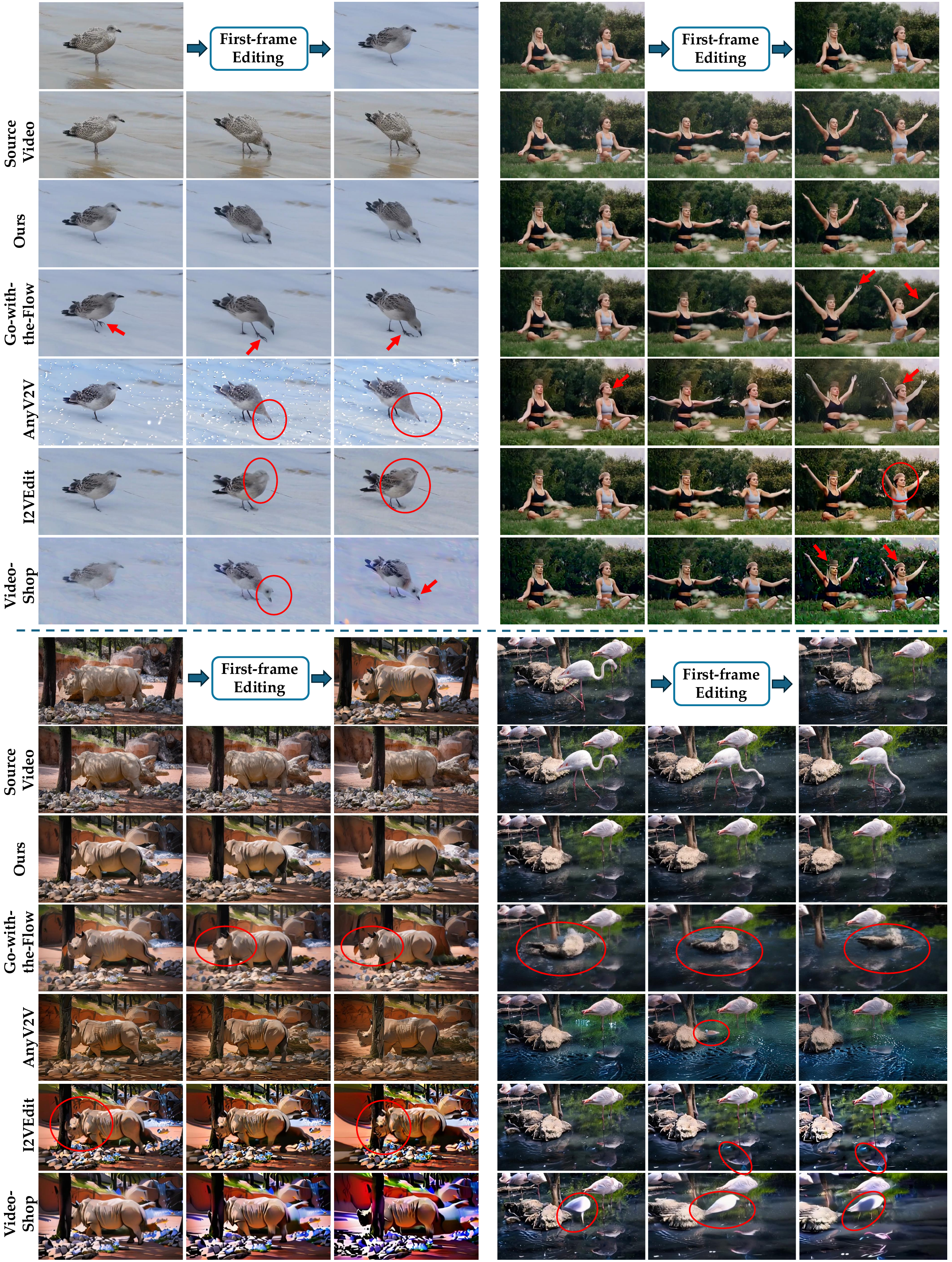}
  \caption{\small Qualitative comparison to existing image-guided video editing methods Go-with-the-Flow~\cite{burgert2025go},  AnyV2V~\cite{anyv2v}, I2VEdit~\cite{i2vedit}, and VideoShop~\cite{videoshop} on various editing scenarios: \textit{adding/removing objects, style transfer, and background editing}. 
}
  \label{fig:qualitative}
    
\end{figure*}

\begin{table*}[t]
\small
\centering
\setlength\tabcolsep{1pt}
\begin{tabular}{l|c|c|c|c|c|c|c|c}
\toprule  
\multirow{2}*{Settings}  
&\multicolumn{2}{c|}{Whole-Image} &\multicolumn{5}{c|}{Non-editing Area}  &\multirow{2}*{Speed (min)}
\\ 
&\textit{CLIP score}           $\uparrow$
&\textit{Warp Error}           $\downarrow$     
&\textit{CLIP score}$^{+}$           $\uparrow$
&\textit{Warp Error}$^{+}$ $\downarrow$ 
&\textit{SSIM}$^{+}$           $\uparrow$   
&\textit{PSNR}$^{+}$           $\uparrow$   
&\textit{LPIPS}$^{+}$ $\downarrow$ 
\\
\hline
AnyV2V~\cite{anyv2v} &0.908   &2.190 &0.911 &1.872 &0.684 &22.08 &0.218 &3.61\\
I2VEdit~\cite{i2vedit}  &0.909   &2.886 &0.914 &1.794  &0.745 &22.57 &0.255 &127\\
VideoShop~\cite{videoshop} &0.822  &7.415  &0.854  &5.298  &0.568  &18.08 &0.417 &2.45\\
Go-with-the-Flow~\cite{burgert2025go} &\underline{0.930}  &{2.152}   &\underline{0.935} & 1.765  & 0.737 &22.66 &0.306 &3.42\\
 \rowcolor{mygray}
FREE-Edit \textit{w/} REE injection  &\textbf{0.937} &\textbf{0.595}  &\textbf{0.941} &\textbf{0.471}  &\textbf{0.811} &\textbf{27.05} &\textbf{0.161} &\underline{1.53}\\
\hline
FREE-Edit \textit{w/} vanilla injection  &0.891 &\underline{1.656} & 0.898 &\underline{1.128}  &\underline{0.782} &\underline{26.14} &\underline{0.202} &1.55\\
FREE-Edit \textit{w/o} injection  &0.873 &13.262  &0.884 & 11.771 &0.586 &20.67  &0.583 &\textbf{1.52}\\
\bottomrule
\end{tabular}
\caption{\small Quantitative  comparison with the existing first-frame-guided video editing methods. We also provide the results of FREE-Edit \textit{w/o} injection, and \textit{w/} vanilla injection. The best and second-best results are marked with bold and underlines.}
\vspace{-0.3cm}
\label{tab:quanti}
\end{table*}

\subsection{Qualitative Results}
In Fig~\ref{fig:quali1}, we first present a qualitative comparison between FREE-Edit \textit{w/o} injection, \textit{w/} vanilla injection, and \textit{w/} REE injection, where the first-frame mask is also listed. FREE-Edit \textit{w/o} injection cannot maintain the video motion well (see movement of people). FREE-Edit \textit{w/} vanilla injection suffers from the issue of undesired content leakage. In contrast, our standard FREE-Edit can achieve the best performance, which demonstrates its capability of preserving the non-editing area while performing effective editing propagation. In the last row of each example, we show the editing mask warped by the first-frame mask, which still corresponds to the editing area well throughout the entire video.

Then, in Fig.~\ref{fig:qualitative}, we present a qualitative comparison with state-of-the-art image-driven video editing methods: Go-with-the-Flow~\cite{burgert2025go}, VideoShop~\cite{videoshop}, AnyV2V~\cite{anyv2v}, and I2VEdit~\cite{i2vedit}. Note that VideoShop and I2VEdit can only process around 14 and 20 frames respectively on an NVIDIA A100 GPU. Following the official implementation of I2VEdit, we split each video into chunks of 20 frames for VideoShop and 14 frames for I2VEdit, where the last edited frame of the previous chunk is used as the initialization for the current chunk.

As shown in Fig.~\ref{fig:qualitative}, I2VEdit~\cite{i2vedit} fails to maintain appearance consistency as the frame increases. The training-free methods, AnyV2V~\cite{anyv2v} and VideoShop~\cite{videoshop}, exhibit more severe degradation in visual quality. The very recent method Go-with-the-Flow~\cite{burgert2025go} can present a good appearance, while facing inaccurate subtle motion (see human arm). In contrast, our FREE-Edit, with the proposed REE injection method, yields the most stable and visually coherent results. Our approach effectively propagates the appearance from the first edited frame to subsequent frames. More examples are provided in the Appendix. 

\begin{table}[t]
\small
\centering
\begin{tabular}{l|c|c}
\toprule  
{Ours \textit{vs}}    
&\makecell[c]{Editing\\Consistency}  $\uparrow$
&\makecell[c]{Overall\\ Quality} $\uparrow$    \\ 
\hline
AnyV2V  & 79.6\% & 81.9\%\\
I2VEdit   &78.5\% &74.6\%\\
VideoShop  & 85.2\% & 81.4\%\\
Go-with-the-Flow & 72.5\% & 68.4\%\\
\hline
FREE-Edit \textit{w/o} injection   &82.2\% & 71.3\%\\
FREE-Edit \textit{w/} vanilla injection  &77.8\% &69.8\%\\
\bottomrule
\end{tabular}
  \vspace{-0.6em}
\caption{\small User Study. Our method outperforms different editing techniques across both aspects.
}
\label{tab:user}
\end{table}


\subsection{Quantitative Results}
In Tab.~\ref{tab:quanti}, we further conduct a quantitative evaluation on the I2V-Edit-Benchmark. We also include results for FREE-Edit \textit{w/o} injection and \textit{w/} vanilla injection. For whole-image measurement, our standard FREE-Edit achieves the highest \textit{CLIP score} and lowest \textit{Warp Error}, demonstrating its superiority in both edit consistency and temporal coherence. Besides, our method achieves the best results on all metrics for non-editing areas. Specifically, as a training-free method, our method has 19.8\% higher PSNR than I2VEdit.

Besides, we record the time consumption of each method over 10 editing trials with 41 frames per edit. Thanks to the real-time performance of LTX-Video-2B~\cite{hacohen2024ltx}, our FREE-Edit achieves the fastest speed. The three variants of our FREE-Edit achieve roughly comparable speeds. Note that I2VEdit requires additional training for motion extraction and takes about \textbf{2 hours} for a 41-frame editing video on an NVIDIA A100 GPU.

We also conduct a user study by applying a standard two-alternative forced-choice format for each participant. Specifically, users were given the source video, the edited first frame, and two editing results, where one is our FREE-Edit and the other is the corresponding competing method. Each user is required to select a better performer regarding \textit{Editing Consistency} ($\uparrow$) and \textit{Overall Quality} ($\uparrow$). We report the selected ratios in Tab.~\ref{tab:user}, where ours achieves a win rate higher than the random chance of 50\% for each competing method, demonstrating our superiority in both editing quality and video generation quality.

\begin{table}[t]
    \centering
    \small
    \begin{tabular}{l|c|c}
    \toprule  
      Method   & \textit{CLIP-Image} $\uparrow$    & \textit{Warp Error} $\downarrow$ \\
    \hline
   TokenFlow~\cite{tokenflow}   & - & 1.777 \\
 \hline
   VidToMe~\cite{li2024vidtome} &-& 1.570 \\
 \hline
   RF-Edit~\cite{wang2024taming} &-& 1.729 \\
 \hline
  \rowcolor{mygray}
   Ours & \textbf{0.937}  & \textbf{0.595} \\
     \bottomrule
    \end{tabular}    
    \caption{\small Quantitative comparison with state-of-the-art text-guided methods: TokenFlow~\cite{tokenflow}, VidToMe~\cite{li2024vidtome}, and RF-Edit~\cite{wang2024taming}.
    }
\label{tab:text_compare}
\end{table}

\subsection{Comparison With Text-guided Methods}
We compare our method with state-of-the-art text-guided methods: TokenFlow~\cite{tokenflow}, VidToMe~\cite{li2024vidtome}, and RF-Edit~\cite{wang2024taming}. As shown in Fig.~\ref{fig:text_driven}, previous methods TokenFlow~\cite{tokenflow} and VidToMe~\cite{li2024vidtome} cannot perform removing or adding objects. Besides, although the very recent method RF-Edit~\cite{wang2024taming} can deal with these two editing scenarios, great changes have also taken place in the faces of two women and the appearance of the remaining two ducks. In contrast, our method can achieve the most consistent editing results.

\begin{figure*}[t]
  \centering
  \includegraphics[width=0.88\linewidth]{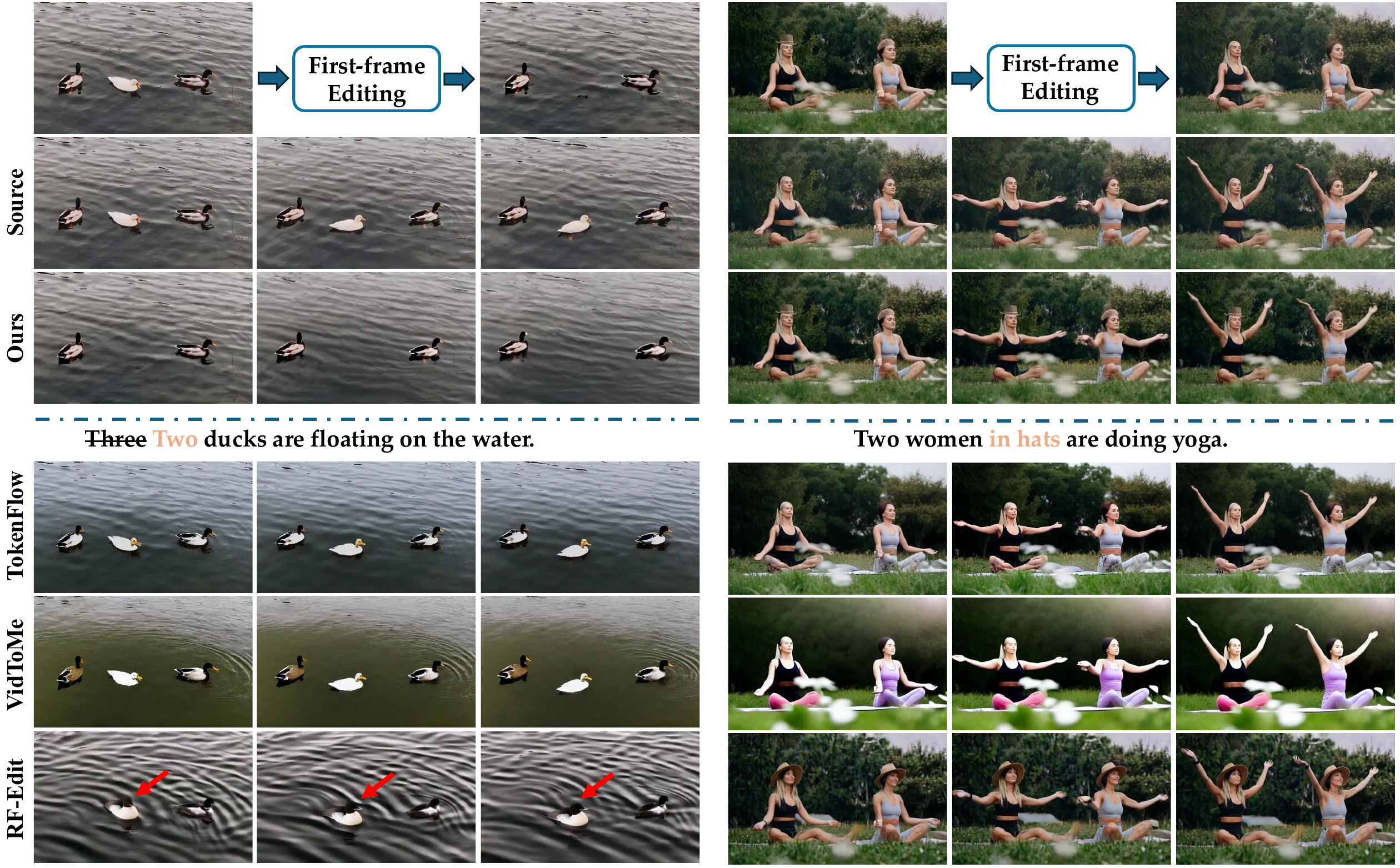}
  \caption{\small Qualitative comparison to existing text-guided video editing methods: TokenFlow~\cite{tokenflow}, VidToMe~\cite{li2024vidtome}, and RF-Edit~\cite{wang2024taming}.
}
\label{fig:text_driven}
\end{figure*}
\begin{figure}[t]
  \centering
  \includegraphics[width=1\linewidth]{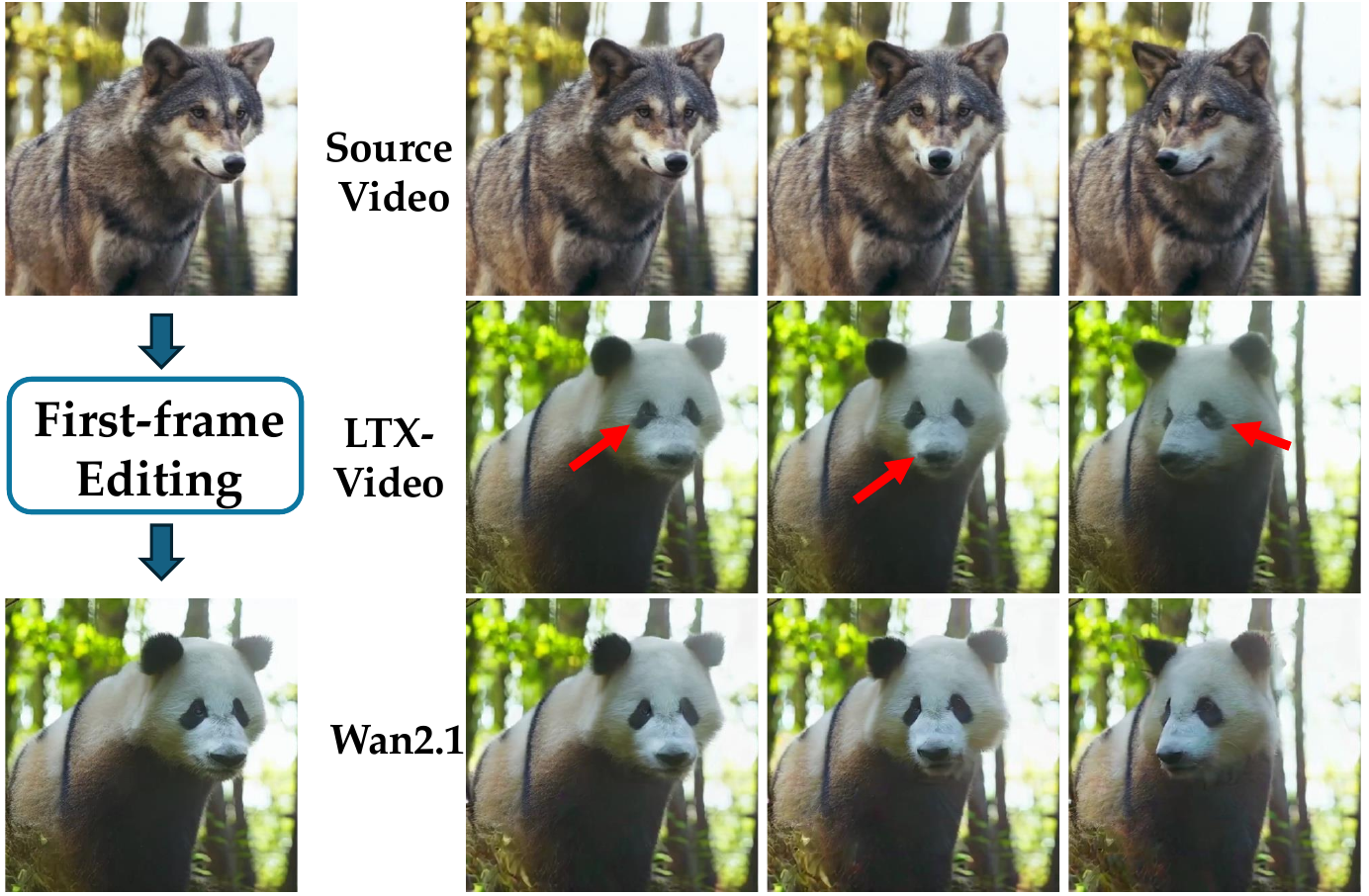}
    \small
  \caption{\small FREE-Edit based on Wan2.1-14B~\cite{wanx} and LTX-Video-2B~\cite{hacohen2024ltx}. Although time-consuming, Wan2.1-14B has the potential to provide better frame quality than LTX-Video-2B.}
  \label{fig:diff_base}
\end{figure}
\begin{figure}[t]
  \centering
  \includegraphics[width=1\linewidth]{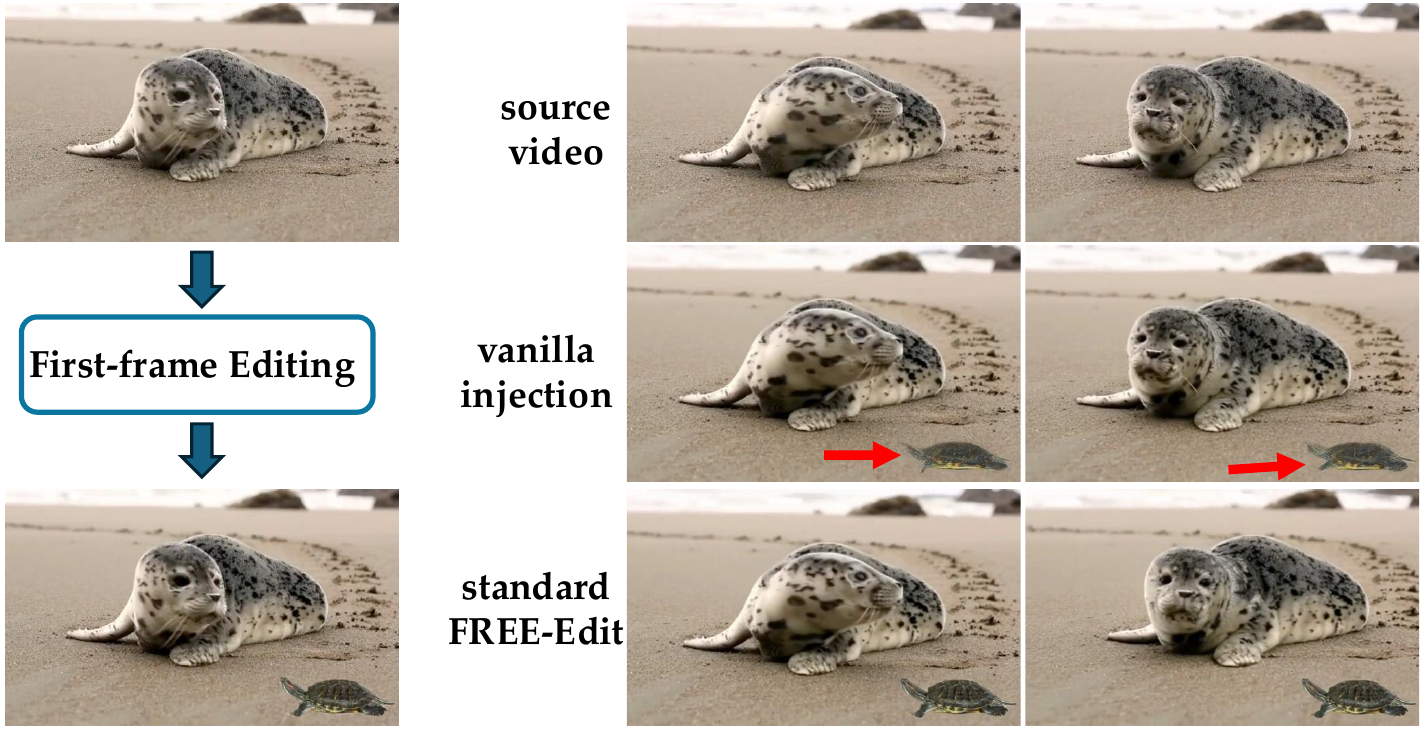}
  \caption{\small Our limitation. When performing \textit{adding objects}, our FREE-Edit cannot endow them with free motion. For example, a newly added turtle cannot have a new trajectory because there is no corresponding motion information in the source video.}
    \label{fig:limitation}
\end{figure}

In Tab.~\ref{tab:text_compare}, we provide a quantitative comparison on \textit{Warp Error}, which shows the advantages of our method in motion consistency. Note that we cannot calculate \textit{CLIP-Image score} for these methods, since they are text-driven.

\subsection{Extension to Different RF models}
Our REE injection can be equipped on different RF models. In this paper, we use LTX-Video-2B~\cite{hacohen2024ltx} as the default I2V model since its real-time generation. Specifically, its speed is $5\sim6$ times faster than Wan2.1-14B~\cite{wanx} while achieving satisfactory results. In Fig.~\ref{fig:diff_base}, we can further observe that although using Wan2.1-14B as the base I2V model is time-consuming, it has the potential to provide results with stronger image quality.

\section{Limitations}
While our REE injection method achieves state-of-the-art performance, it has several limitations. Most notably, it cannot generate free motion for newly added objects. As shown in Fig.~\ref{fig:limitation}, although our method maintains better appearance consistency compared to vanilla injection, it cannot produce natural motion for inserted objects (\eg, turtle). This is also a limitation of existing methods. We leave it as future work by incorporating trajectories of new objects.


\section{Conclusion}
In this paper, we propose an \textbf{E}diting-awa\textbf{RE} (REE) feature injection method, which designs a modulation weight $\bm{\lambda}$ to adaptively assign feature injection weights for each token. Specifically, we first use optical flow to warp the automatically calculated first-frame editing mask, and thus obtain editing masks for subsequent frames. Based on them, we compute the modulation weight $\bm{\lambda}$ to encourage that no injection is performed in editing areas. Further, we propose FREE-Edit, which integrates our REE injection method with a pre-trained \textbf{F}low model for image-driven video editing. Extensive experiments have demonstrated the effectiveness of our method.

\newpage

{
    \small
    \bibliographystyle{ieeenat_fullname}
    \bibliography{main}
}

\clearpage
\clearpage
\setcounter{page}{1}
\setcounter{figure}{0}
\setcounter{table}{0}
\renewcommand{\figurename}{\textbf{Ap-Fig.}}
\renewcommand{\tablename}{\textbf{Ap-Tab.}}

\appendix

\twocolumn[{%
\renewcommand\twocolumn[1][]{#1}%
\maketitlesupplementary
\begin{center}
    \centering
    \captionsetup{type=figure}
\includegraphics[width=1.0\textwidth]{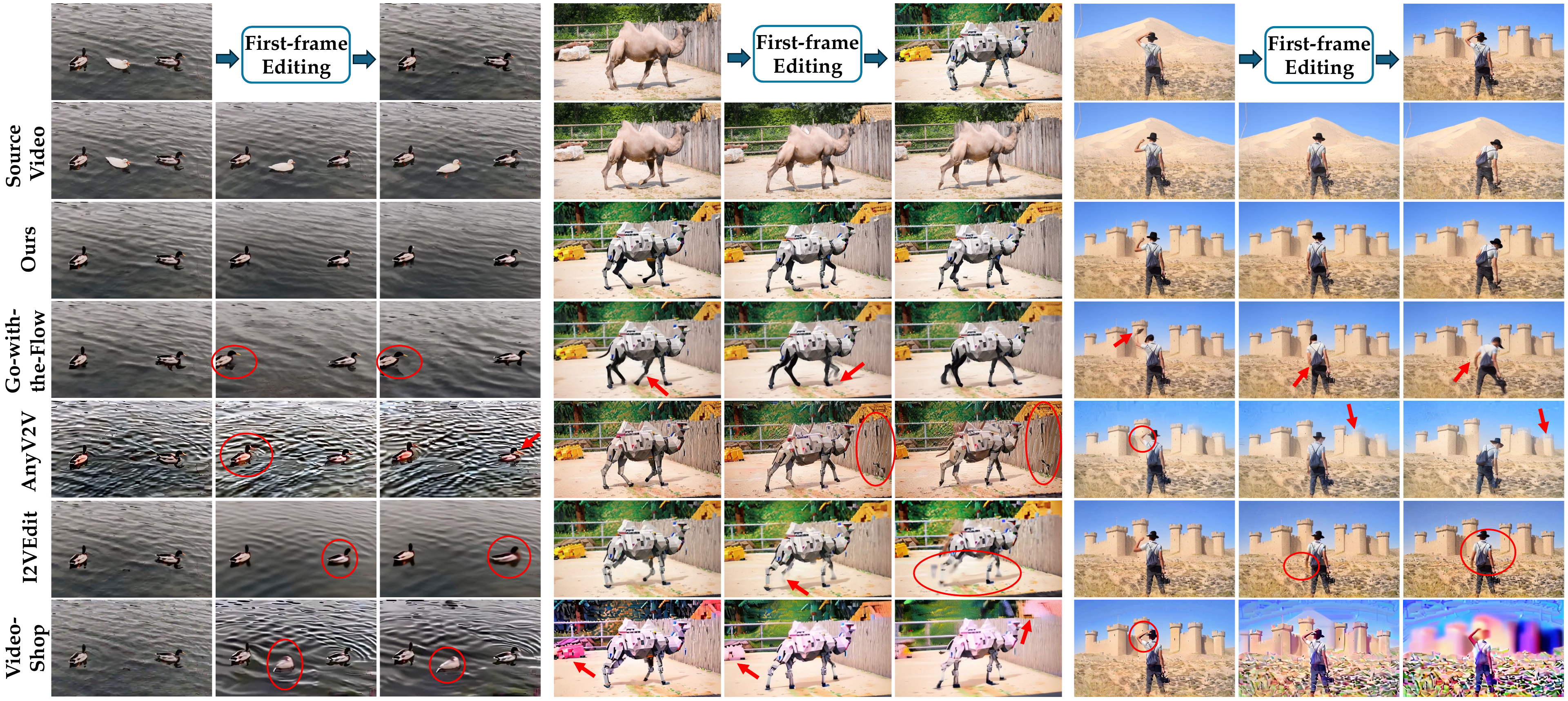}
  \caption{\small More comparison to existing image-guided video editing methods: Go-with-the-Flow~\cite{burgert2025go},  AnyV2V~\cite{anyv2v}, I2VEdit~\cite{i2vedit}, and VideoShop~\cite{videoshop} on various editing scenarios: \textit{removing objects, style transfer, and background editing}. }
  \label{fig:sm_quali}
\end{center}%
}]


\section{More Visualization Results}
Recall that in Fig. 4 in our main paper, we present a qualitative comparison between FREE-Edit \textit{w/o} injection, \textit{w/} vanilla injection, and \textit{w/} REE injection on \textit{adding object} and \textit{object replacement}. In this Appendix, we provide an additional example on \textit{style transfer}. Ap-Fig~\ref{fig:sm-ree} shows that FREE-Edit \textit{w/o} injection struggles to preserve motion fidelity in the source video. Even in the style transfer case, vanilla injection tends to encounter unwanted semantic leakage (see the nose and ears of the rhino). In the last row, we show the editing masks warped by the first-frame mask.

\begin{figure}[t]
  \centering
  \includegraphics[width=0.8\linewidth]{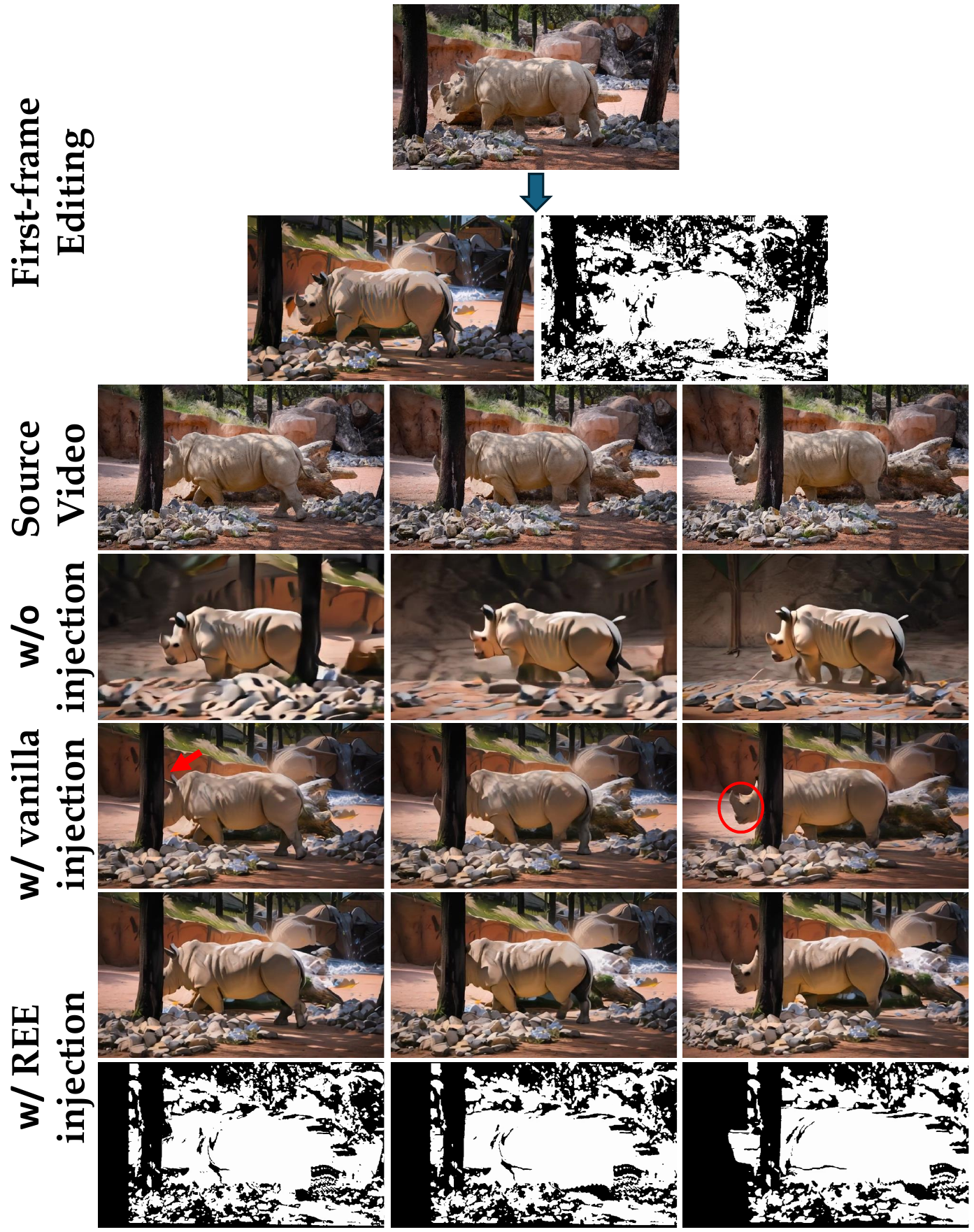}
  \caption{\small More comparison between standard FREE-Edit (\textit{w/} REE injection), and that of \textit{w/o} injection, and \textit{w/} vanilla injection. }
  \label{fig:sm-ree}
\end{figure}

\begin{figure*}[t]
  \centering
  \includegraphics[width=0.88\linewidth]{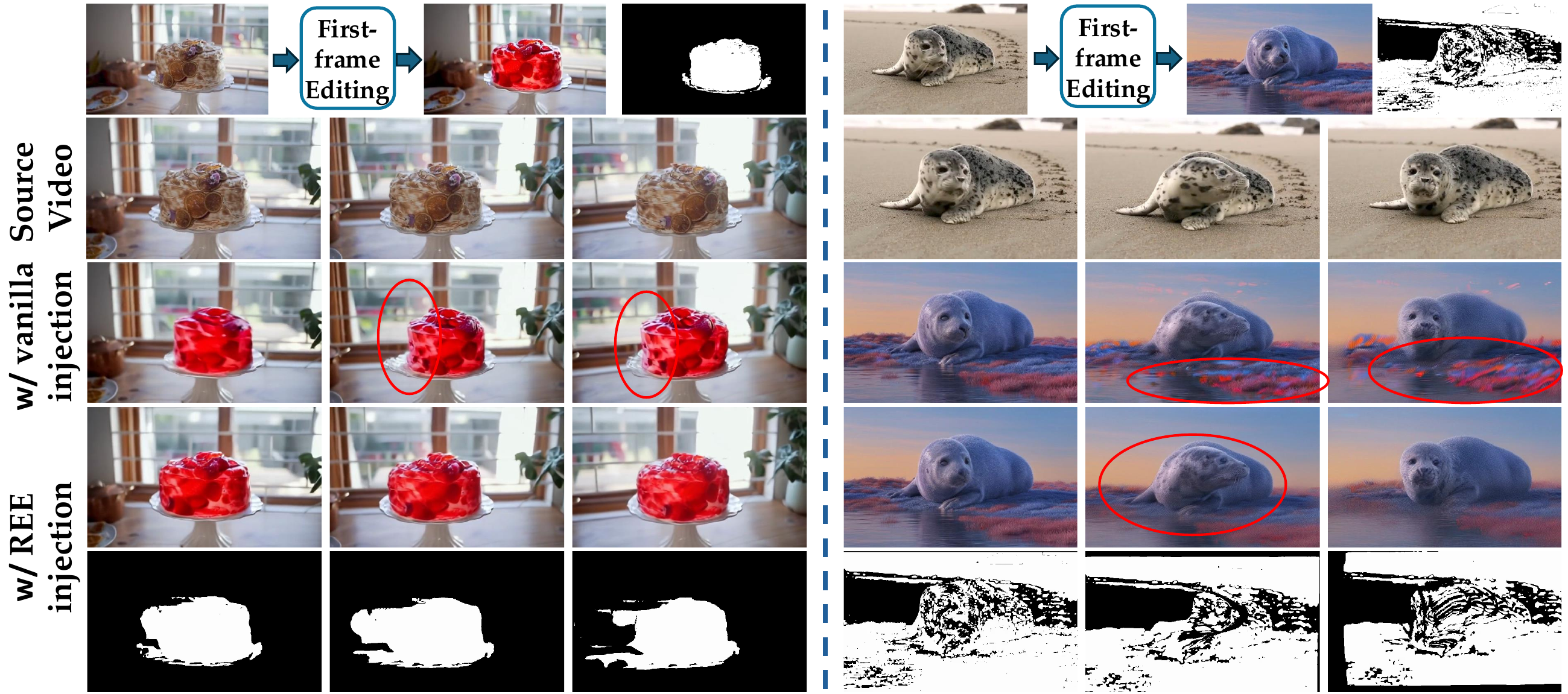}
  \caption{ \small Left: Results of large changes in viewpoint. Right: A failure case with fast motion.}
  \label{fig:optical}
\end{figure*}

Besides, in Ap-Fig.~\ref{fig:sm_quali}, we show more qualitative comparison with the current state-of-the-art image-guided video editing methods, including Go-with-the-Flow~\cite{burgert2025go}, VideoShop~\cite{videoshop}, AnyV2V~\cite{anyv2v}, and I2VEdit~\cite{i2vedit}. Our FREE-Edit demonstrates significant superiority over existing approaches. Compared to I2VEdit~\cite{i2vedit}, which is a method that requires one-shot training for each example, our approach exhibits better appearance and motion consistency as the frame number increases. Moreover, Go-with-the-Flow~\cite{burgert2025go} always suffers from the problem of inaccurate motion.

\section{Discussion on Optical Flow Accuracy}
Ap-Fig.~\ref{fig:optical} presents complex examples with rotating camera views (lemon cake → strawberry cake) and fast motion (turning head and returning). Our REE injection can successfully propagate semantics during large viewpoint changes, whereas vanilla injection results in an incorrect cake shape. Here we also provide the results of editing-mask propagation. It can be seen that inaccurate optical flow may appear at object edges. FREE-Edit alleviates its influence to a certain degree using the prior of the pretrained I2V model.

We also provide a failure case under fast motion, where vanilla injection yields unsatisfactory background and object appearance.
Although our REE injection can generate a background consistent with the edited first frame, it cannot fully correct object appearance due to inaccurate optical flow. We argue that a stronger optical flow model can improve the performance of our FREE-Edit.

\end{document}